\documentclass[10pt,twocolumn,letterpaper]{article}

\usepackage[pagenumbers]{cvpr} 

\makeatletter
\@namedef{ver@everyshi.sty}{}
\makeatother

\usepackage{graphicx}
\usepackage{amsmath}
\usepackage{amssymb}
\usepackage{booktabs}
\usepackage{gensymb}

\DeclareMathOperator*{\argmin}{arg\,min}
\usepackage{colortbl}
\usepackage{multirow}
\usepackage{svg}
\usepackage{multicol}
\usepackage{makecell}
\usepackage{comment}

\definecolor{mygreen}{rgb}{0.0, 0.8, 0.0}
\newcommand{\todo}[1]{\textcolor{blue}{{[#1]}}}

\usepackage{hyperref}
\hypersetup{colorlinks,allcolors=blue}

\usepackage[capitalize]{cleveref}
\crefname{section}{Sec.}{Secs.}
\Crefname{section}{Section}{Sections}
\Crefname{table}{Table}{Tables}
\crefname{table}{Tab.}{Tabs.}

\def\cvprPaperID{6733} 
\def\confName{CVPR}
\def\confYear{2023}

\begin{document}

\title{Rotation-averaging Should Include Two-View Uncertainties}
\title{Uncertainty-Aware Rotation Averaging}
\title{Secrets of Rotation Averaging}
\title{Revisiting Rotation Averaging: Uncertainties and Robust Losses}

\author{Ganlin Zhang\\
ETH Zürich\\
{\tt\small zhangga@student.ethz.ch}
\and
Viktor Larsson\\
Lund University\\
{\tt\small viktor.larsson@math.lth.se}
\and
Daniel Barath\\
ETH Zürich\\
{\tt\small dbarath@inf.ethz.ch}
}

\twocolumn[{
\maketitle
\vspace{-1em}
    \centering
    \small
    \setlength{\tabcolsep}{8pt}
    \newcommand{\sz}{0.3}
    \begin{tabular}{cc}
        \includegraphics[width=0.48\linewidth,trim={0 0 0 0},clip]{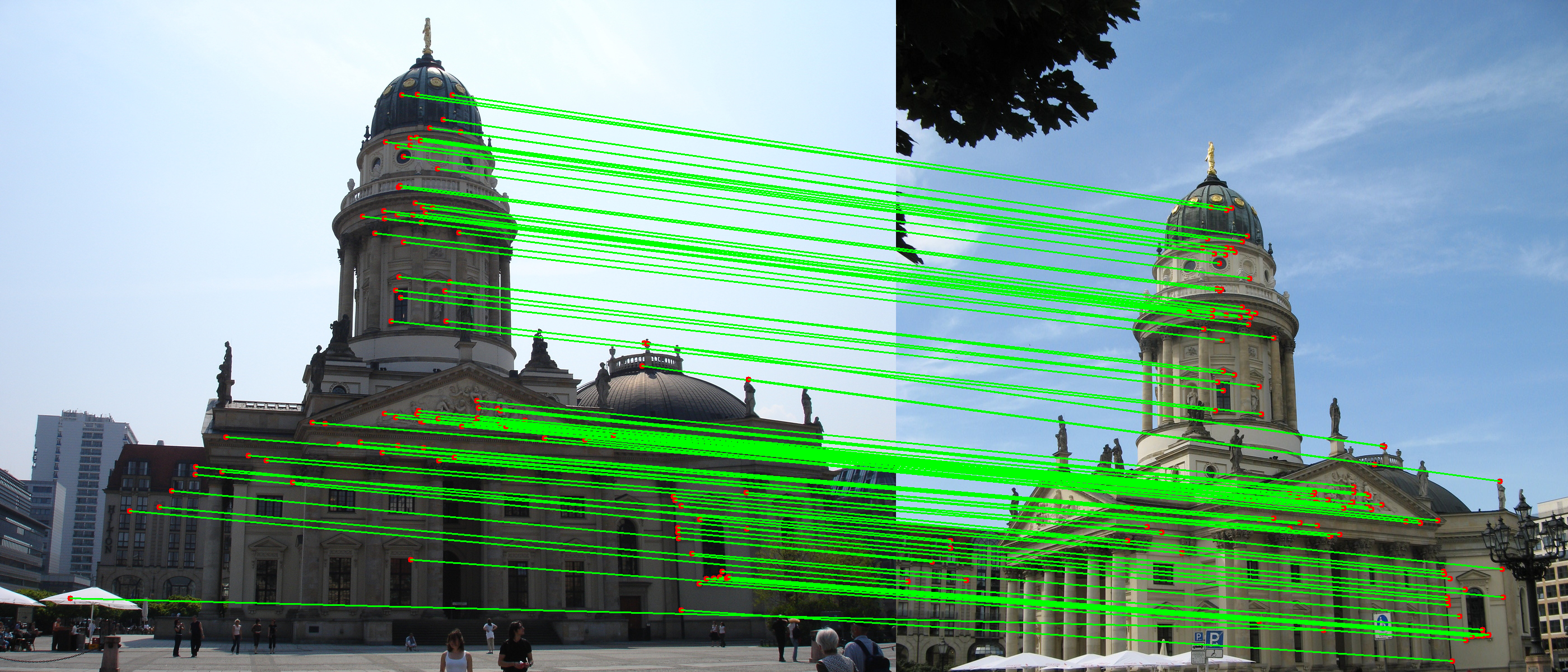} & 
        \includegraphics[width=0.48\linewidth,trim={40mm 0 100mm 0},clip]{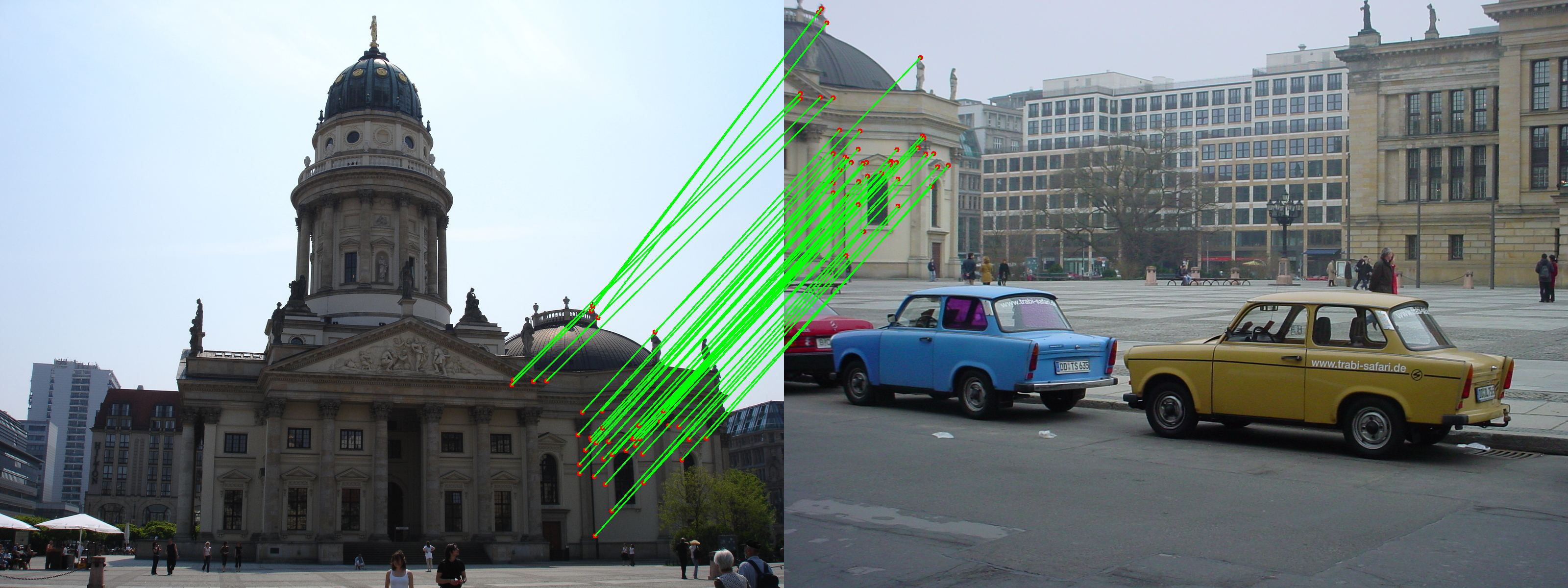} \\
        \multirow{2}{*}{\makecell{(a) Uncertainty = \textcolor{mygreen}{\textbf{0.63}}\\
        Rotation error = \textcolor{mygreen}{\textbf{0.05\degree}}, Translation error = \textcolor{mygreen}{\textbf{6.30\degree}}}} & 
        \multirow{2}{*}{\makecell{(b) Uncertainty = \textcolor{red}{\textbf{7.61}}\\
        Rotation error = \textcolor{red}{\textbf{4.81\degree}}, Translation error = \textcolor{red}{\textbf{21.12\degree}}}} \\
         & \\[-5pt]
    \end{tabular}
    \captionof{figure}{
      Two example pairs where, even though the relative pose estimation found a large number of correct inliers (shown by green line segments), the quality of the estimated poses differs greatly.  
      The uncertainties, used as weighting terms, help in rotation averaging to account for this quality difference. 
      Here, the shown numbers as the uncertainties are the traces of the covariance matrices.}
    \label{fig:teaser}
    \vspace{15pt}
}]

\begin{abstract}
 In this paper, we revisit the rotation averaging problem applied in global Structure-from-Motion pipelines.
 We argue that the main problem of current methods is the minimized cost function that is only weakly connected with the input data via the estimated epipolar geometries.
 We propose to better model the underlying noise distributions by directly propagating the uncertainty from the point correspondences into the rotation averaging.
 Such uncertainties are obtained for free by considering the Jacobians of two-view refinements. 
 Moreover, we explore integrating a variant of the MAGSAC loss into the rotation averaging problem, instead of using classical robust losses employed in current frameworks. 
 The proposed method leads to results superior to baselines, in terms of accuracy, on large-scale public benchmarks. 
 The code is public. \href{https://github.com/zhangganlin/GlobalSfMpy}{https://github.com/zhangganlin/GlobalSfMpy} 
\end{abstract}

\section{Introduction}
\label{sec:intro}

 Building large 3D reconstructions from unordered image collections is an essential component in any system that relies on crowd-sourced mapping.
The current paradigm is to perform this reconstruction via \textit{Structure-from-Motion}\cite{schonberger2016structure} which jointly estimates the camera parameters and the scene geometry represented with a 3D point cloud.
Methods for Structure-from-Motion can generally be categorized into two classes;
\textit{Incremental methods} \cite{snavely2006photo,snavely2008modeling,wu2013towards,schonberger2016structure} that sequentially grows a seed reconstruction by alternating triangulation and registering new images, and
\textit{Global methods} \cite{martinec2007robust,olsson2011stable,moulon2013global,cui2015global} which first estimate pairwise geometries and then aggregate them in a bottom-up approach.
 Historically, incremental methods are more robust and accurate, but the need for frequent bundle adjustment \cite{triggs1999bundle} comes with significant computational cost which limits their scalability.
 In contrast, global (or non-sequential) methods require much lower computational effort and can in principle scale to larger image collections.
However, in practice, current methods are held back by the lack of accuracy and have not enjoyed the same level of success as incremental methods.

Global methods work by first estimating a set of pairwise epipolar geometries between co-visible images.
Next, via \textit{rotation averaging}, a set of globally consistent rotations are estimated by ensuring they agree with the pairwise relative rotations.
Once the rotations are known,  the camera positions and 3D structure are estimated, and refined jointly in a single final bundle adjustment.

Rotation averaging has a long history in computer vision (see \eg~\cite{govindu2001combining, martinec2007robust} for early works) and is a well-studied problem.
Most methods formulate it as an optimization problem, finding the rotation assignment that minimizes some energy.
A common choice is the \textit{chordal distance}, measuring the discrepancy in the rotation matrices in the $L_2$-sense
\begin{equation} \label{eq:chordal}
     \min_{\{R_i\}_{i=1}^N} \sum_{i=1}^N \sum_{j\in\mathcal{N}(i)} \| \hat{R}_{ij} R_i - R_j \|_F^2
\end{equation}
where $\hat{R}_{ij}$ is the relative rotation estimated between image $i$ and $j$.
There are also other choices such as using angle-axis~\cite{sweeney2015theia} or quaternion~\cite{govindu2001combining} as rotation representation, or optimising over a Lie algebra~\cite{govindu2004lie}, however the overall idea (measuring some consistency with the relative estimates) remains the same.
 Many works have focused on the optimization problem itself, both theoretically \cite{wilson2016rotations,eriksson2019rotation} and by providing new algorithms \cite{dellaert2020shonan}, but did not consider whether the cost itself is suitable for the task.
 In \eqref{eq:chordal}, each relative rotation measurement is given the same weight. However in practice, the quality of the epipolar geometries varies significantly. Figure~\ref{fig:teaser} shows two images with wildly different uncertainties (and errors) in the rotation estimate. 
  To address this problem, there is a line of work \cite{hartley2011l1,chatterjee2013efficient,sidhartha2021all} which augment the cost in \eqref{eq:chordal} with robust loss functions that give a lower weight to large residuals.
 However, the same loss function is generally applied to each residual, independent of the measurement uncertainty.
 
\begin{figure}[t]
  \centering
  \includegraphics[width=1\linewidth]{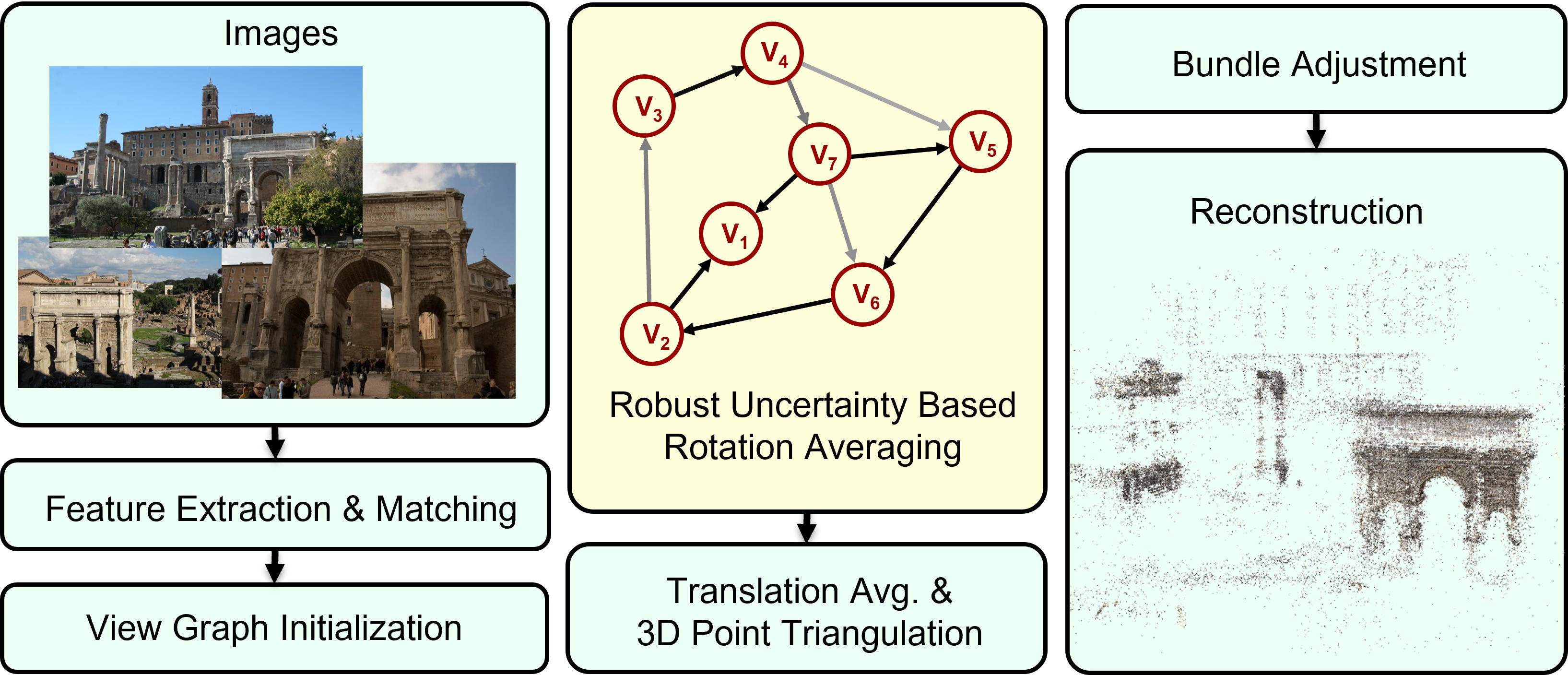}
  \caption{\textbf{Global Structure from Motion pipeline} with the proposed uncertainty-based rotation averaging. In the yellow box, the opacity of the view graph edges indicates uncertainty difference.}
  \label{fig:pipeline}
\end{figure}

 In this paper we revisit the rotation averaging problem.
 We argue that the main problem in current methods is that the cost functions that are minimized are only weakly connected with the input data via the estimated epipolar geometries.
We propose to better model the underlying noise distributions (coming from the keypoint detection noise and spatial distribution) by directly propagating the uncertainty from the point correspondences into the rotation averaging problem, as shown in Figure~\ref{fig:pipeline}.
While the idea itself is simple, we show that this allows us to get significantly more accurate estimates of the absolute rotations; reducing the gap between incremental and global methods.
Note that the uncertainties we leverage are essentially obtained for free by considering the Jacobians of the two-view refinement.

As a second contribution, we explore integrating a variant of the MAGSAC~\cite{barath2021marginalizing} loss into the rotation averaging problem, instead of using the classical robust losses employed in current frameworks. 
MAGSAC\cite{barath2021marginalizing} was originally proposed as a threshold-free estimator for two-view epipolar geometry, where the idea is to marginalize over an interval of acceptable thresholds, \ie, noise range. 
We show that this fits well into the context of rotation-averaging, as it is not obvious how to set the threshold for deciding on inlier/outlier relative rotation measurements, especially in the uncertainty-reweighted cost that we propose.

\section{Related Work}
\textbf{Rotation averaging} is a long standing problem in computer vision with some of the early works dating back more than two decades. Govindu~\cite{govindu2001combining}  proposes an approximation in which the problem becomes linear in terms of the quaternions, after heuristically resolving the sign ambiguity. Similarly, Martinec and Pajdla~\cite{martinec2007robust} parameterize the problem in terms of the full $3\times 3$ matrix, but drop the non-linear constraints to obtain a tractable problem. Wilson et al. \cite{wilson2016rotations} investigate, more generally, under what conditions the rotation averaging problem is easy.

When the chordal distance is used, \ie~\eqref{eq:chordal}, the optimization problem can be solved via an SDP-based relaxation, see Arie et al.~\cite{arie2012global} and Fredriksson et al.~\cite{fredriksson2012simultaneous} for some of the earlier papers in this line of work. 
For this particular relaxation, Eriksson et al.~\cite{eriksson2019rotation} show that under some noise assumptions, the SDP relaxation obtains the globally optimal solution. In~\cite{eriksson2019rotation}, the authors propose a block-coordinate descent method specialized for the dual formulation of the rotation averaging problem. 
Dellaert et al.~\cite{dellaert2020shonan} propose an optimization scheme based on sequentially lifting the problem into higher-dimensional rotations $\text{SO}(n)$. 
The method, named \textit{Shonan} rotation averaging, is shown to avoid some local minima in which standard optimization techniques, such as Levenberg-Marquardt~\cite{more1978levenberg}, might be stuck in. 
In our work, our contributions are related to changing the cost function minimized, and it is possible that the methods from these works could be applied in our setting as well.

To deal with outliers in the relative rotation measurements, Hartley et al.\cite{hartley2011l1} propose a generalization of the Weiszfeld-algorithm to minimize the $L_1$-loss over the rotation residuals. 
This method was later extended by Chatterjee and Govindu in \cite{chatterjee2013efficient}. 
To obtain more robust estimations other robust losses have been explored, \eg~$L_{0.5}$ \cite{chatterjee2017robust} and Geman-McClure~\cite{sidhartha2021all}. 
In our work, we experimentally evaluate these loss-functions (in addition to many others~\cite{huber1992robust,holland1977robust}) and compare against the loss function based on MAGSAC~\cite{barath2021marginalizing} that we propose. In~\cite{gao2021incremental}, Gao et al.~propose an iterative scheme for solving the rotation averaging problem where they weight the view graph edges based on the two-view inliers. 
In our experiments, we compare against this re-weighting scheme as well.

\textbf{Global Structure-from-Motion} builds the reconstruction by aggregating pair-wise estimates of epipolar geometries. 
In most cases, this is done via rotation averaging~\cite{olsson2011stable,moulon2013global}, but there are also works that perform the averaging in $\text{SE}(3)$ instead~\cite{cui2015global}. Once rotations are known, there are different approaches for recovering the translations and structure. 
Wilson and Snavely~\cite{wilson_eccv2014_1dsfm} solve the translation averaging problem using an outlier filter based on projecting the translations onto 1D subspaces. 
Moulon et al.~\cite{moulon2013global} formulates the problem as an $L_\infty$-optimization. 
Olsson and Enqvist~\cite{olsson2011stable} also rely on  $L_\infty$-optimization but solve jointly for both 3D points and camera positions.

There are several open-source frameworks for global Structure-from-Motion, such as Theia~\cite{sweeney2015theia} and OpenMVG~\cite{moulon2016openmvg}. 
In our experiments, we integrate our updated optimization objectives into the framework from Theia~\cite{sweeney2015theia}, and use the remaining pipeline unchanged. 
Our contributions are, however, not specific to this pipeline.

\section{Rotation Averaging with Uncertainties}
\label{sec:rot_uncertainty}

In this paper, we propose a way to leverage uncertainties, coming from pair-wise relative pose estimations, in rotation averaging.  
This additional signal indicating the quality of the input relative poses allows to further improve 3D reconstruction by global Structure from Motion (SfM) methods. 

\begin{figure}[t]
  \centering
  \includegraphics[width=0.9\linewidth,trim={50mm 23mm 50mm 20mm},clip]{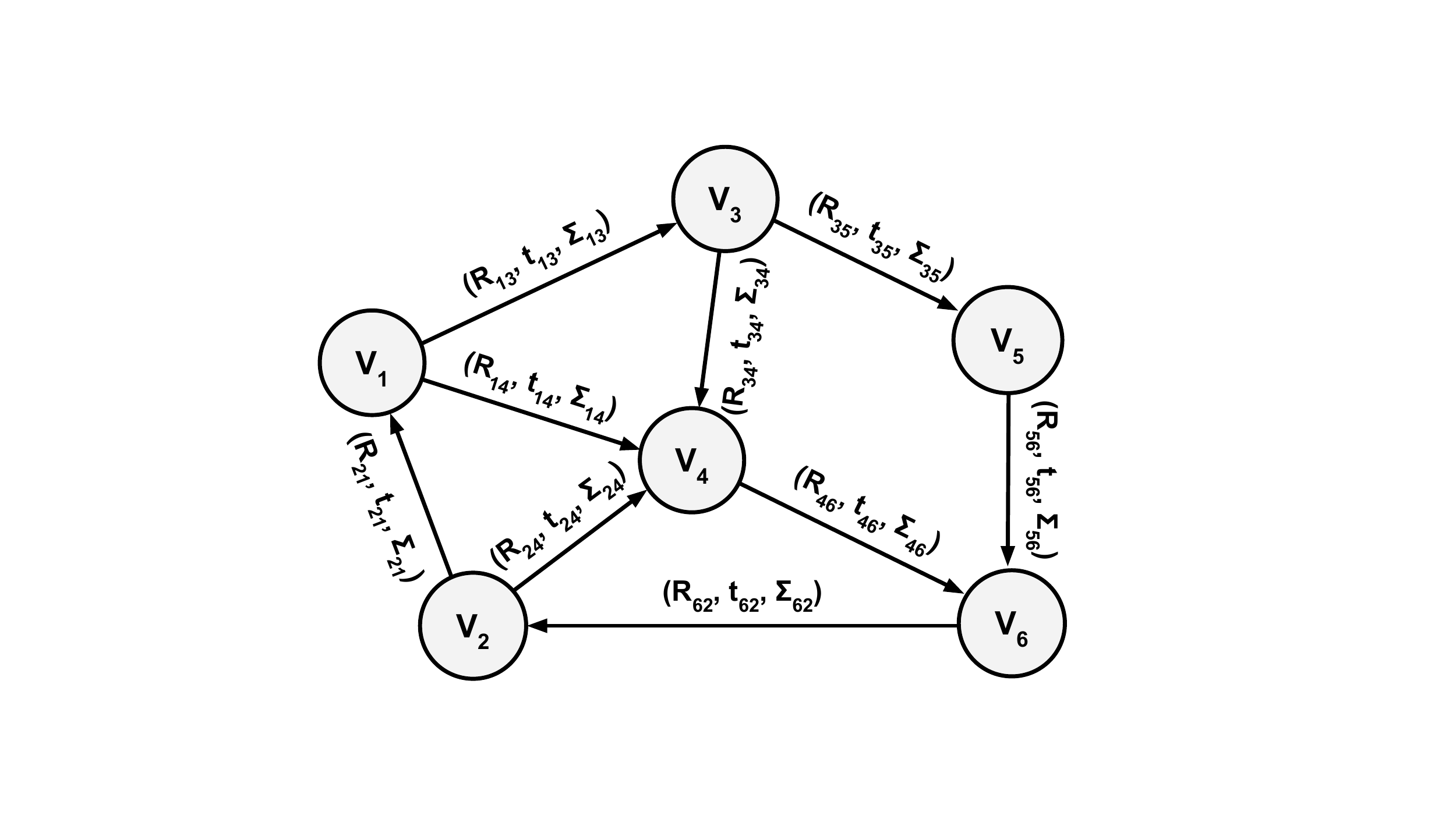}
  \caption{\textbf{View graph} with its vertices representing images and its edges 3D rotations and translations between the views.
  Edge weight $\Sigma_{ij}$ comes from the uncertainty of the pair-wise estimation.}
  \label{fig:viewgraph}
\end{figure}

\paragraph{Overview of Rotation Averaging.}
Rotation averaging is one of the main components of global SfM algorithms allowing to estimate the global rotation in such a way that it is decoupled from the translations. 
It is usually formalized as a graph optimization problem, where the vertices are the image poses and the edges are relative poses estimated as a preliminary step. See Fig.~\ref{fig:viewgraph}.
Each edge imposes a constraint on the global poses of its neighboring vertices (\ie, images). 
The primary goal is to use these constraints to find the absolute poses of the cameras.

Standard rotation averaging proceeds as follows: 
first, a vertex is chosen as the origin. 
Second, the orientations of all vertices that fall into the same connected component as the origin-defining one are initialized by a maximum spanning tree.
Finally, the rotations are optimized jointly, leveraging the information coming from the relative rotations. 
Rotation $R_i^*$ of the $i$th camera is calculated as 
\begin{equation}\label{eq:overview_rot_avg}
\begin{aligned}
    \{R^*_i\}_{i=1}^n = \argmin_{\{R_i\}_{i=1}^n} \sum_{(i,j)\in \mathcal{E}} \rho\left(\left\|L_e(R_i,R_j,R_{ij})\right\|^2\right) , 
\end{aligned}
\end{equation}
where $\mathcal{E}$ is the set of edges in the view graph $\mathcal{G} = (\mathcal{E}, \mathcal{V})$,
function $L_e$ measures the error of the relative rotation $\bar{R}_{ij} = R_i R_j ^\text{T}$ coming from the optimized global orientations $R_i$ and $R_j$ of the $i$th and $j$th views w.r.t.\ the estimated rotation $R_{ij}$.
Function $\rho(\cdot) : \mathbb{R} \to \mathbb{R}$ is the robust loss used to deal with potential outliers in the data.
The optimization procedure outputs  $\{R^*_i\}_{i=1}^n$, the estimated absolute orientations of all the $n$ views in the same connected component.
In case multiple disjoint graphs are generated from the images, the procedure runs on all of them independently. 

\paragraph{Uncertainty-Aware Rotation Averaging.}

In \cref{eq:overview_rot_avg}, every edge $(i,j) \in \mathcal{E}$ of the view graph $\mathcal{G}$ is treated equally. 
However, the estimated relative poses depend on many factors in practice (\eg, inlier number, baseline) and thus, they are of different quality.
This quality measure can be captured by the uncertainty of the two-view estimation. 
In rotation averaging, the uncertainty in each measurement can be considered in a weighting scheme as follows:
%
%
\begin{equation}
    \small
    \{R^*_i\}_{i=1}^n = \argmin_{\{R_i\}_{i=1}^n}\sum_{(i,j)\in E} \rho\left( \left\|D_{ij}^\text{T} L_e(R_i,R_j,R_{ij})\right\|^2\right),
\label{eq:weighted_rot_avg}
\end{equation}
where term $D_{ij}$ depicts the uncertainty of edge $(i, j)$.
%

\begin{figure*}
  \centering
  \begin{subfigure}{0.49\linewidth}
    \includegraphics[width=0.94\linewidth]{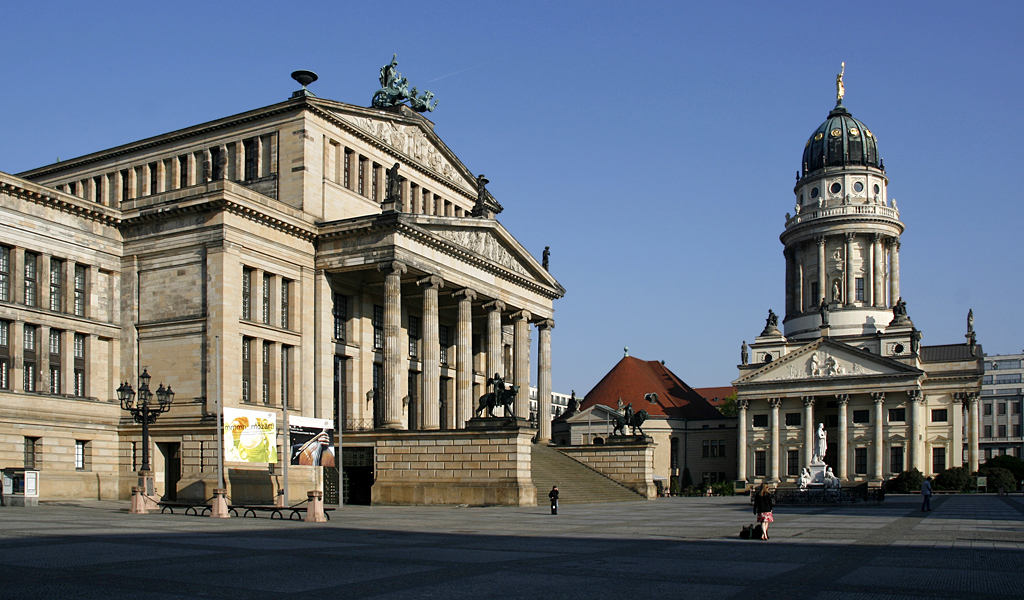}
    \caption{Example image from scene Gendarmenmarkt.}
    \label{fig:gdm-img}
  \end{subfigure}
  \begin{subfigure}{0.49\linewidth}
    \includegraphics[width=0.94\linewidth]{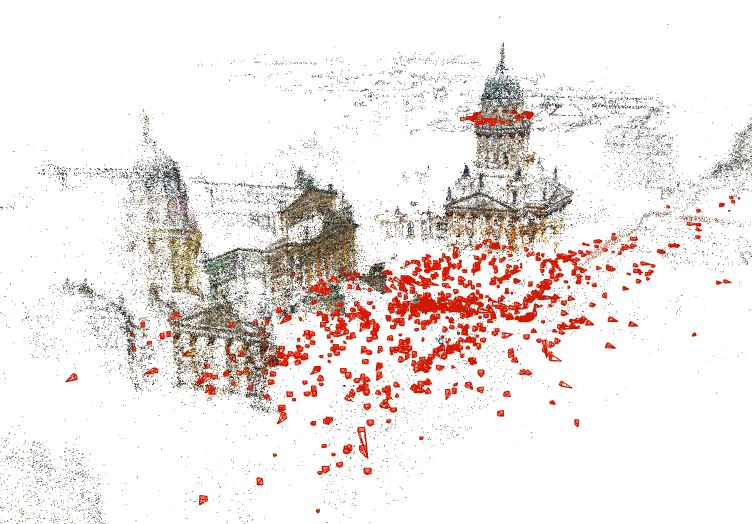}
    \caption{Ground truth.}
    \label{fig:gdm-gt}
  \end{subfigure}
  \begin{subfigure}{0.49\linewidth}
    \includegraphics[width=0.94\linewidth]{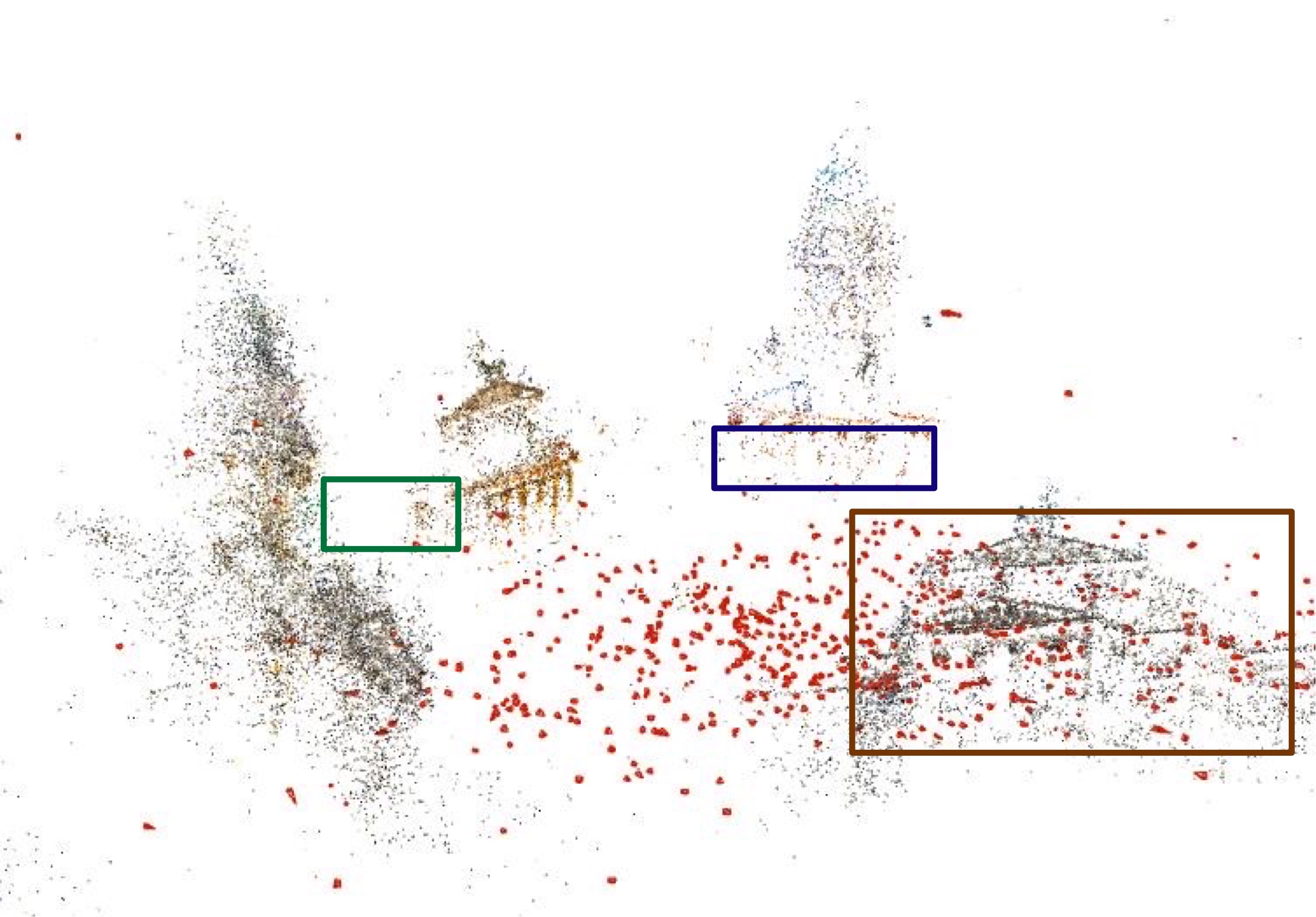}
    \caption{Baseline.}
    \label{fig:gdm-baselin}
  \end{subfigure}
  \begin{subfigure}{0.49\linewidth}
    \includegraphics[width=0.94\linewidth]{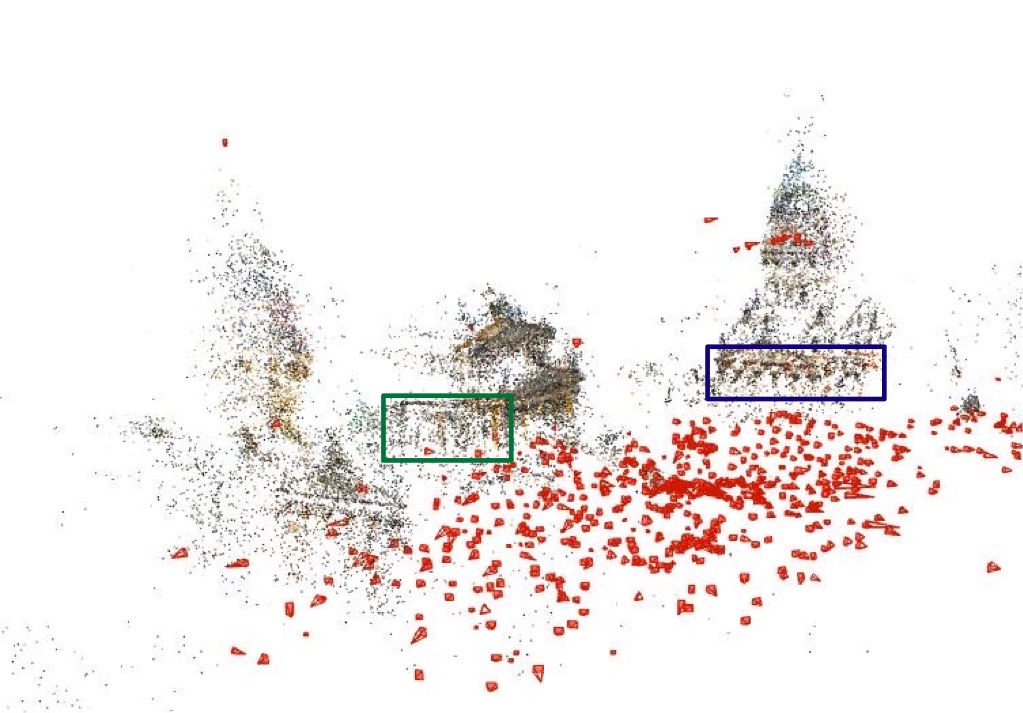}
    \caption{Ours.}
    \label{fig:gdm-ours}
  \end{subfigure}
  \caption{
  Scene Gendarmenmarkt from the 1DSfM dataset, containing a concert hall in the middle and two cathedrals on both sides. 
  \cref{fig:gdm-img} is an image from the scene. 
  \cref{fig:gdm-gt} is the ground truth reconstruction. 
  \cref{fig:gdm-baselin} and \cref{fig:gdm-ours} are results of the baseline~\cite{sweeney2015theia} and the proposed method, respectively. 
  The camera poses are drawn as red cones. 
  Compared to the baseline, our method has achieve better reconstruction. 
  Example differences are shown by colored boxes in \cref{fig:gdm-baselin} and \cref{fig:gdm-ours}.
  \textit{Green boxes}: walls of the concert hall are missing in the baseline, while being reconstructed in ours. 
  \textit{Blue boxes}: the bottom of the right cathedral is almost missing in the baseline, but is well-preserved in ours. 
  \textit{Brown box}: baseline reconstructs an additional concert hall which does not exist. 
  \cref{fig:gdm-ours} shows that the proposed method is correct. 
  }
  \label{fig:short}
\end{figure*}

\paragraph{Relative Rotation Uncertainty.} 

To propagate the uncertainty from the features to the estimated relative rotation, we first express the rotation via the 2D-2D point correspondences. 
The relative rotation $R_{ij}$ and translation $t_{ij}$ between the $i$th and $j$th views is written as follows:
\begin{equation}\label{eq:uncertainty_loss}
\begin{aligned}
R_{ij},t_{ij} = &\argmin_{R,t}L_{ij} \\  
= &\argmin_{R,t}\sum_{(p,p')\in \mathcal{M}_{ij}}{L_s}^2(R,t,p,p',K_i,K_j),
\end{aligned}
\end{equation}
where $\mathcal{M}_{ij} = \left\{ (p,p') \; | \; p \in \text{view}_{i} \wedge p' \in  \text{view}_{j}\right\}$ is the set of point correspondences in views $i$ and $j$. 
Matrices $K_i$ and $K_j$ contain the intrinsic parameters, \eg focal length and principal point, of the cameras. 
$L_s$ is the point-to-model residual, \eg, Sampson distance or symmetric epipolar error.
Due to its robustness, we use Sampson distance as $L_s$ written as follows:
\begin{eqnarray}\label{eq:uncertainty_sampson}
    L_s(R_{ij},t_{ij},p,p',K_i,K_j) = \nonumber\\ 
 \dfrac{p^\text{T}{}' F_{ij} p }{\left((Fp)_1^2+(Fp)_2^2+(F^\text{T} p')_1^2+(F^\text{T} p')_2^2\right)^{1/2}},
\end{eqnarray}
where $F_{ij}$ is the fundamental matrix as
\begin{equation}
    F_{ij} = {K_j}^{-\text{T}} R_{ij} [t_{ij}]_\times {K_i}^{-1},
\end{equation}
and the lower-indices refer to coordinates of $F p$ and $F^\text{T} p'$.
%

The uncertainty propagation to get covariance $C(R_{ij})$ of relative rotation $R_{ij}$ is done as
\begin{equation}\label{eq:uncertainty_prop}
    C(R_{ij}) = (J^\text{T}(R_{ij})J(R_{ij}))^{-1}
\end{equation}
where $J(R_{ij})$ is the Jacobian of $L_{ij}$. Here axis-angle parameterization of rotation matrix is used to avoid singularity.

Covariances are plugged into Eq.~\ref{eq:weighted_rot_avg} by calculating $D_{ij}$ as the LLT decomposition of $C(R_{ij})^{-1}$~\cite{nocedal1999numerical} that equals to 
\begin{equation}
    C(R_{ij})^{-1} = D_{ij}D_{ij}^\text{T}.
\end{equation}
Using such weighting scheme in the optimization procedure allows us to consider the pair-wise uncertainties of the rotations to reason about their qualities in a theoretically justifiable manner.
In practice, $L_s$, and thus Jacobian $J(R_{ij})$, is calculated only on the inlier correspondences after the robust estimation, \eg RANSAC~\cite{fischler1981random}, finishes.

\subsection{Marginalizing over the Noise Scale}\label{sec:magsac_loss}

In MAGSAC++~\cite{barath2021marginalizing}, a robust loss function is designed by marginalizing over the noise standard deviation to reduce the dependence on a manually set inlier-outlier threshold parameter. 
The MAGSAC++ loss assumes that the inlier residuals follow a $\chi^2$-distribution and does not make any assumptions about the outliers.
While the MAGSAC++ loss has only been applied to residual functions measuring the consistence of point correspondences with a projective transformation (\eg, homography and essential matrix), it can be used as a general robust loss like the Huber loss~\cite{huber1992robust}.

Let us define the MAGSAC++ loss for rotation averaging.
The data points, in our case, are a set of relative rotations, and the models to estimate are the global orientations of the cameras. 
This means that for a pair of global rotations $R_i$ and $R_j$, we are given relative rotation $R_{ij}$ constraining both.
The inlier weight of $R_{ij}$ is 
\begin{equation}\label{eq:magsacweight}
    \small
    w(r(R_i R_j^\text{T}, R_{ij})) = \int_{0}^{+\infty} \mathbf{P}(R_{ij} \; | \; R_i, R_j, \sigma) f(\sigma) \text{d}\sigma,
\end{equation}
where $r(R_i R_j^\text{T}, R_{ij})$ is the residual, $\sigma$ is the noise standard deviation, $f(\sigma)$ is the prior distribution of $\sigma$ assumed to be uniform on range $[0, \sigma_{\max}]$.
%

By rewriting \cref{eq:magsacweight} as the marginal density of the inlier residual, we can get
\begin{equation}\label{eq:magsacweight2}
    w(r) = \int_{0}^{+\infty}g(r \; | \; \sigma) f(\sigma) \text{d}\sigma,
\end{equation}
where $g(r\;|\;\sigma)$ is the density of the residual $r$ given $\sigma$.
Assume that $\sigma$ is uniformly distributed $\sigma \sim U(0,\sigma_\text{max})$,
then \cref{eq:magsacweight2} becomes
\begin{equation}\label{eq:magsacweight3}
    w(r) = \frac{1}{\sigma_\text{max}}\int_{0}^{\sigma_\text{max}}g(r \; | \; \sigma)\text{d}\sigma.
\end{equation}

Assume that the residual $r$ is in some $\nu$-dimensional space and the error along each axis of this $\nu$-dimensional space is independent and normally distributed with the same variance $\sigma^2$. 
Then $r^2/\sigma^2$ has $\chi^2$-distribution with $\nu$ degrees of freedom. For a given $\sigma$, $r$ has the trimmed $\chi$-distribution with $\nu$ degrees of freedom. Let $\tau(\sigma)=k\sigma$ be the chosen quantile of the $\chi$-distribution. 

For $r\geq k\sigma_\text{max}$, $w(r)=0$. For $0\leq r\leq k\sigma_\text{max}$
\begin{equation}\label{magsacweight4}
\begin{aligned}
    w(r) &= \frac{1}{\sigma_\text{max}} \int_{r/k}^{\sigma_\text{max}}g(r\;|\;\sigma)\text{d}\sigma= \\ 
    \frac{1}{\sigma_\text{max}} M(\nu)2^{\frac{\nu-1}{2}} &\left(\Gamma\left(\frac{\nu-1}{2},\frac{r^2}{2\sigma_\text{max}^2}\right)-\Gamma\left(\frac{\nu-1}{2},\frac{k^2}{2}\right) \right).
\end{aligned}
\end{equation}
Here, $M$ is the normalizing constant as follows:
\begin{equation}
    M(\nu) = \left(2^{\frac{\nu}{2}} \Gamma\left(\frac{\nu}{2}\alpha\right)\right)^{-1},
\end{equation}
where $\alpha$ means that $\tau(\sigma)$ is set to the $\alpha$-quantile, \ie
$\text{CDF}(k)=\alpha$. 
The CDF here is the cumulative distribution function of $\chi$-distribution with $\nu$ degrees of freedom.
In our case, $\nu = 3$. 
Function $ \Gamma(a,x) $ is the upper-incomplete gamma function 
\begin{equation}
    \Gamma(a,x) = \int_{x}^{+\infty}t^{a-1}\exp{(-t)}\text{d}t,
\end{equation}
and 
\begin{equation}
    \Gamma(a) = \Gamma(a,0) = \int_{0}^{+\infty}t^{a-1}\exp{(-t)}\text{d}t.
\end{equation}

From \cref{magsacweight4}, we can calculate the weight of each edge by passing its residual to the formula. 
In order to convert it to a robust loss that can be used in the optimization procedure while keeping its beneficial properties, we use loss 
%
%
%
\begin{equation}
    \rho(r) = w(0)-w(r),
\end{equation}
where $w(0)$ is the weight if $R_{ij} = R_i R_j^\text{T}$.
It is the maximal weight.
Scalar $w(r)$ is the robust weight that residual $r$ implies. 
Loss $\rho(r)$ is positive and increasing on interval $(0, k \sigma_{\max})$.
Therefore, it can be minimized by IRLS and each iteration
guarantees a non-increase in the loss ({\cite{maronna2019robust}}, chapter
9). Consequently, it converges to a local minimum. 

\section{Experiments}

In this section, we test the global SfM implemented in the Theia~\cite{sweeney2015theia} library with and without considering the uncertainties coming from the estimated relative poses in the proposed way.
Moreover, we evaluate popular loss functions, including Soft L$_{1}$ loss which is used in Theia.
For rotation averaging, 
Theia uses angle-axis rotation parameterization to minimize the relative rotation error  
via a numerical optimization implemented in the Ceres~\cite{Agarwal_Ceres_Solver_2022} library.
It then performs the 1DSfM translation averaging~\cite{wilson_eccv2014_1dsfm}.

\begin{table*}[htbp] 
\centering

\setlength\extrarowheight{0.25pt}
\begin{tabular}{ r |  c c c |c c >{\columncolor[rgb]{0.95,0.95,0.95}}c}
\toprule

  & \multicolumn{3}{c|}{Soft L$_{1}$~\cite{charbonnier1997deterministic}}  & \multicolumn{3}{c}{MAGSAC~\cite{barath2021marginalizing}}\\
  & Baseline & + Inliers & + Covariance & Baseline & + Inliers & + Covariance \\
\hline

Ellis Island 
&68.2&70.0&67.1&73.8&\textbf{76.4}&\underline{76.3}\\
Gendarmenmarkt 
&8.9&7.0&6.1&49.6&\underline{54.0}&\textbf{54.2}\\
Montreal Notre Dame 
&77.2&76.4&74.0&\underline{79.3}&78.6&\textbf{79.3}\\
Notre Dame 
&77.5&\textbf{80.2}&73.9&\underline{80.1}&78.7&79.5\\
NYC Library 
&59.0&61.7&60.6&63.7&\textbf{68.6}&\underline{65.4}\\
Piazza del Popolo 
&60.2&59.3&\underline{62.1}&60.3&\textbf{62.2}&60.7\\
Roman Forum 
&57.7&50.8&60.3&62.5&\underline{65.5}&\textbf{70.2}\\ 
Tower of London			
&48.8&49.4&\underline{66.8}&51.9&57.8&\textbf{67.2}\\
Union Square
&27.2&24.6&31.3&28.0&\textbf{38.0}&\underline{35.0}\\
Yorkminster 
&63.3&64.0&\underline{64.5}&62.3&64.1&\textbf{67.1}\\
Vienna Cathedral
&\textbf{67.1}&\underline{66.8}&60.6&62.9&46.8&66.7\\
Piccadilly
&33.2&33.6&30.6&46.7&\underline{50.0}&\textbf{51.6}\\
Alamo
&63.3&65.4&62.4&\underline{65.4}&65.4&\textbf{66.8}\\
\hline
Average &54.7 &54.6 &55.4&60.5&\underline{62.0}&\textbf{64.6}\\
\bottomrule
\end{tabular}
\caption{The Area Under the recall Curve (AUC) at 5\degree of estimated rotations after rotation averaging by~\cite{chatterjee2013efficient} (Baseline) with different losses (Soft L$_{1}$~\cite{charbonnier1997deterministic} and MAGSAC~\cite{barath2021marginalizing}) and weighting strategies: by the number of inliers (+ Inliers), by the proposed covariance (+ Covariance).
%
}
\label{tab:ra_1dsfm}
\end{table*}

\begin{table}[htbp] 
\centering
\setlength\extrarowheight{0.25pt}
\setlength{\tabcolsep}{1.2mm}
\begin{tabular}{c c c c c c }
\toprule
\multicolumn{2}{c}{\multirow{2}*{Setting}}
&\multicolumn{4}{c}{AUC $\alpha$ (\%) $\uparrow$}\\
 \cmidrule(lr){3-6}
\multicolumn{2}{c}{}& $\alpha$=2\degree & $\alpha$=5\degree & $\alpha$=10\degree & $\alpha$=20\degree\\
\midrule
\multirow{3}*{\shortstack[c]{Soft L$_{1}$\\ \cite{charbonnier1997deterministic}}}
&Baseline & 37.4&	59.4&	72.5&	82.0\\
&+ Inliers &35.9&	58.9&	72.3&	81.9\\
&+ Covariance &39.6&	60.3&	71.5&	79.2\\
\midrule
\multirow{3}*{\shortstack[c]{MAGSAC\\\cite{barath2021marginalizing}}}
&Baseline&41.1&	65.0&	77.4&	85.6\\
&+ Inliers &39.2&	62.3&	74.9&	83.3\\ 
&\cellcolor[rgb]{.95,.95,.95} + Covariance & \cellcolor[rgb]{.95,.95,.95}\textbf{44.6}&	\cellcolor[rgb]{.95,.95,.95}\textbf{67.6}&	\cellcolor[rgb]{.95,.95,.95}\textbf{79.4}&	\cellcolor[rgb]{.95,.95,.95}\textbf{87.2}\\
\bottomrule
\end{tabular}
\caption{AUC scores of rotation errors after the full global SfM pipeline~\cite{sweeney2015theia} averaged over all scenes of the 1DSfM dataset~\cite{wilson_eccv2014_1dsfm}.}
\label{tab:fp_1dsfm}
\end{table}

\begin{figure}
    \centering
    \includegraphics[width=0.482\linewidth]{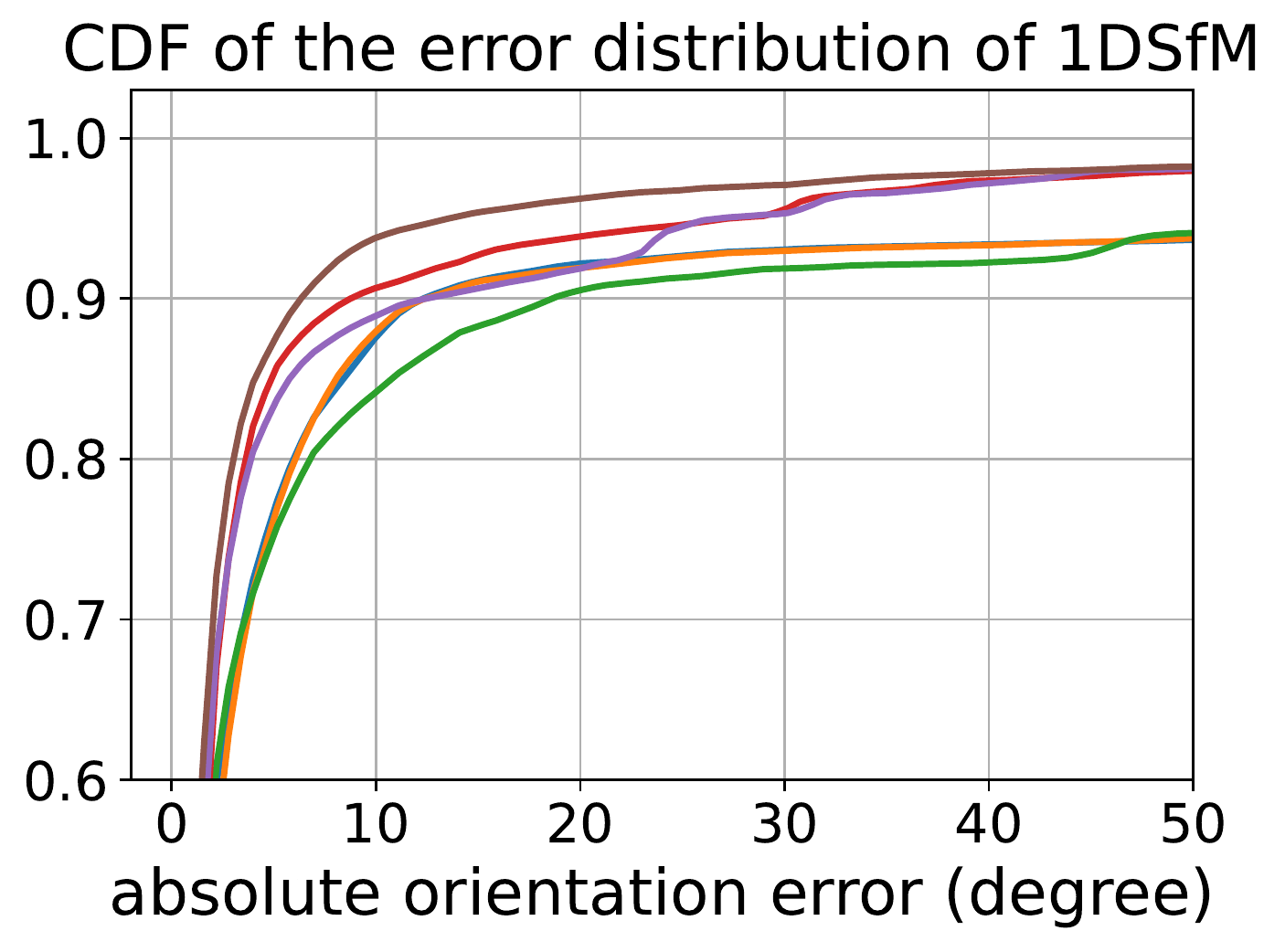}
    \includegraphics[width=0.499\linewidth]{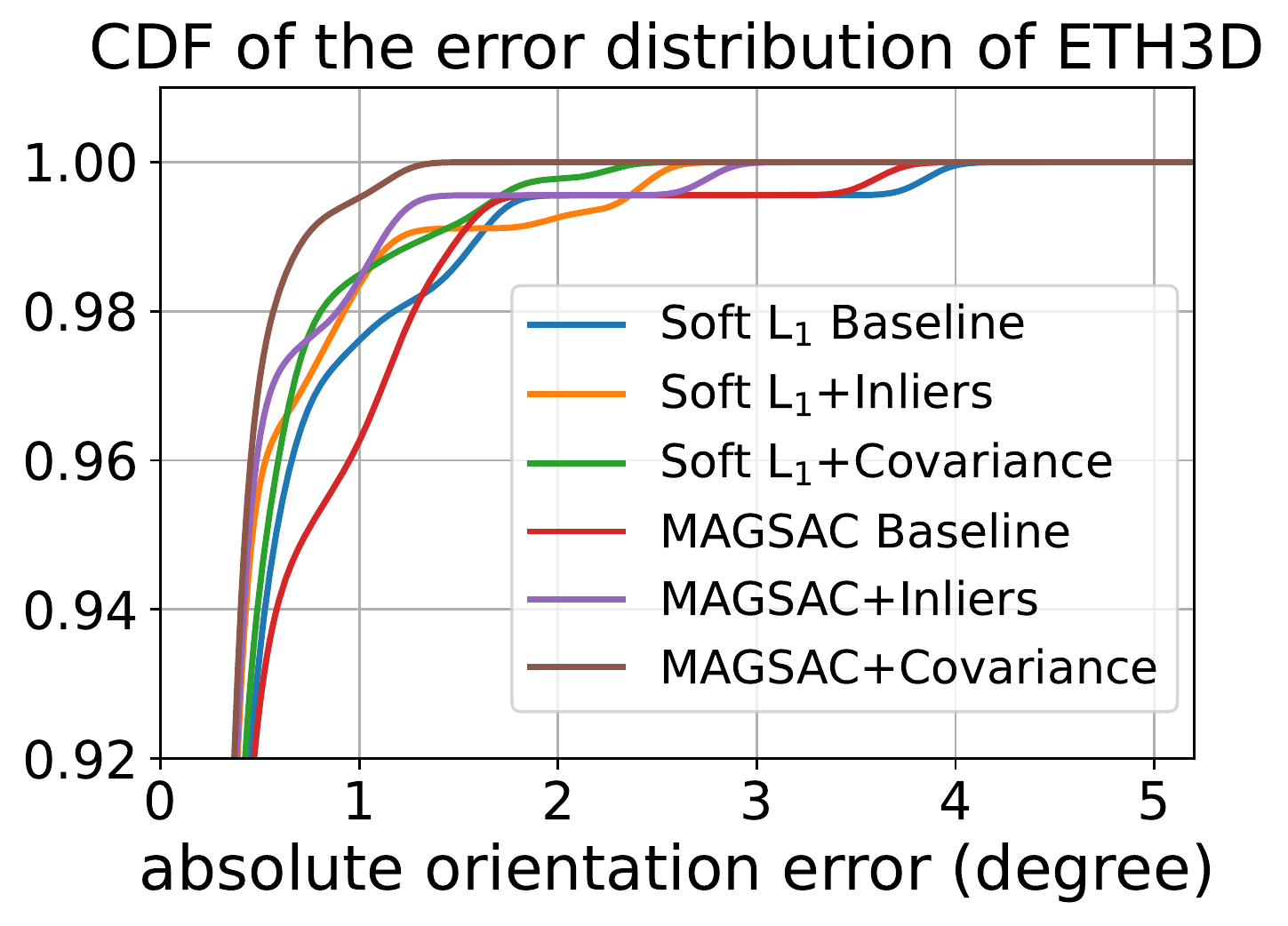}
    \caption{
    Cumulative distribution functions of the rotation errors (in degrees) after the rotation averaging on the 1DSfM~\cite{wilson_eccv2014_1dsfm} (left; 8824 poses) and ETH3D datasets~\cite{schops2017multi} (right; 451 poses). Being close to the top-left corner is preferred.}
    \label{fig:cdfs}
\end{figure}


\begin{figure*}[htbp]
  \centering
  \begin{subfigure}{0.24\linewidth}
    \includegraphics[width=0.98\linewidth]{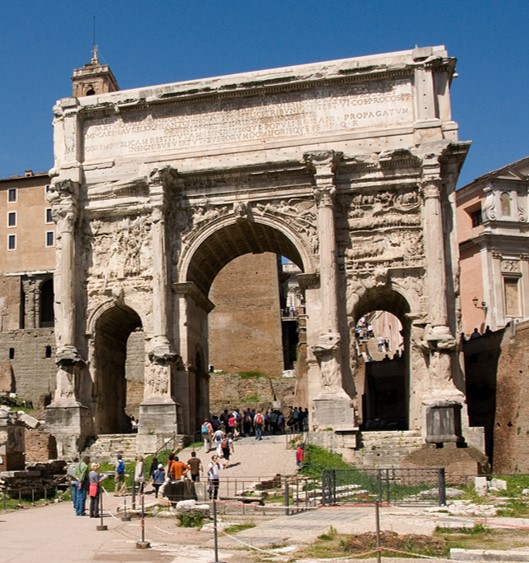}
    \caption{Image of Scene \textit{Roman Forum}.}
    \label{fig:relief-img}
  \end{subfigure}
  \hspace{3pt}
  \begin{subfigure}{0.24\linewidth}
    \includegraphics[width=0.94\linewidth]{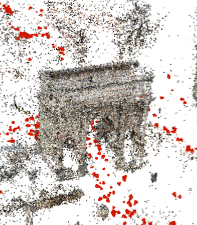}
    \caption{Ground truth.}
    \label{fig:relief-ref}
  \end{subfigure}
  \begin{subfigure}{0.24\linewidth}
    \includegraphics[width=0.94\linewidth]{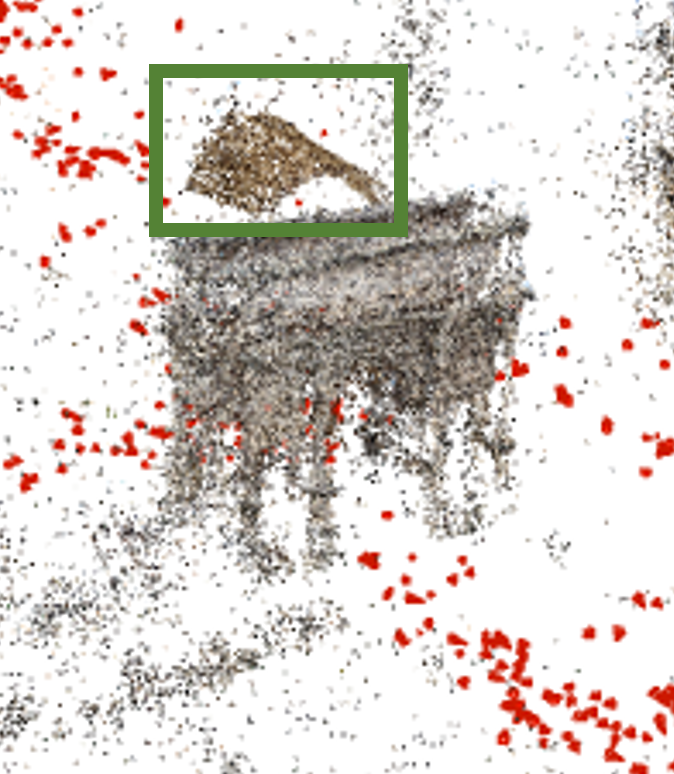}
    \caption{Inliers~\cite{gao2021incremental}.}
    \label{fig:relief-baseilne}
  \end{subfigure}
  \begin{subfigure}{0.24\linewidth}
    \includegraphics[width=0.94\linewidth]{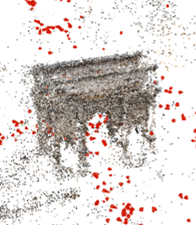}
    \caption{Covariance.}
    \label{fig:relief-ours}
  \end{subfigure}
  \caption{Scene \textit{Roman Forum} from the 1DSfM dataset~\cite{schops2017multi}. \cref{fig:relief-img} is an example image. 
  \cref{fig:relief-ref} is the ground truth reconstruction. 
  \cref{fig:relief-baseilne} and \cref{fig:relief-ours} are the reconstructions 
  when the pose graph edges are weighted by the inlier number~\cite{gao2021incremental} or by the proposed covariance, respectively.
  The cameras are represented by red cones. 
  \textit{Green box}: In \cref{fig:gdm-baselin}, \cite{gao2021incremental} reconstructs an additional roof which does not exist. 
  }
  \label{fig:relief}
\end{figure*}

\paragraph{Benchmarks.}
We test the proposed uncertainty-based rotation averaging on the 1DSfM~\cite{wilson_eccv2014_1dsfm} and ETH3D~\cite{schops2017multi} datasets. 
The 1DSfM dataset consists of 14 different landmarks with images, collected from the Internet, of varying sizes and capturing conditions, \eg, day and night. 
It provides 2-view matches with epipolar geometries and a reference reconstruction from incremental SfM (computed with Bundler~\cite{snavely2006photo,snavely2008modeling}) for measuring error. 
Since Bundler was published more than ten years ago, we reconstructed the scenes with COLMAP~\cite{schonberger2016structure} to get a better reconstruction that can be considered as ground truth. 
We tune the hyper-parameters of the tested methods on scene  Madrid Metropolis, which are the inlier-outlier threshold of loss function in different settings: baseline 0.02, baseline+inlier 0.06, baseline+covariance 0.02.
Therefore, we only report results on the other scenes. 

The ETH3D Stereo Muti-view is an indoor-outdoor dataset of 13 scenes with high-resolution images (6000$\times$4000), keypoints, LiDAR depth, and ground truth poses. 
We use the 13 scenes of the training set, and use all image pairs with at least 500 GT keypoints in common.
For this dataset, we run SuperPoint~\cite{detone2018superpoint} and SuperGlue~\cite{sarlin2020superglue} to obtain point correspondences from which the relative poses are estimated.
As baseline, we also run COLMAP on exactly the same features. 

\begin{table}[htbp] 
\setlength\extrarowheight{0.25pt}
\setlength{\tabcolsep}{1.3mm}
\begin{tabular}{c c c c c c }
\toprule
\multicolumn{2}{c}{\multirow{2}*{Setting}}
&\multicolumn{4}{c}{AUC $\alpha$ (\%) $\uparrow$}\\
 \cmidrule(lr){3-6}
\multicolumn{2}{c}{}& $\alpha$=2\degree & $\alpha$=5\degree & $\alpha$=10\degree & $\alpha$=20\degree\\
\midrule
\multicolumn{2}{c}{COLMAP~\cite{schonberger2016structure}} & 86.7 & 90.9 & 92.8	& 93.8\\
\midrule
\multirow{3}*{\shortstack[c]{Soft L$_{1}$\\ \cite{charbonnier1997deterministic}}}
&Baseline & 88.6&	95.2&	97.6&	\textbf{98.8}\\
&+ Inliers & 89.2&	95.2&	97.3&	98.8\\
&+ Covariance& 84.5&	91.4&	95.6&	97.6\\
\midrule
\multirow{3}*{\shortstack[c]{MAGSAC\\ \cite{barath2021marginalizing}}}
&Baseline   & 86.9&	94.5&	97.3&	98.6\\
&+ Inliers & 89.5&	94.9&	96.9&	98.3\\ 
&\cellcolor[rgb]{.95,.95,.95} + Covariance & \cellcolor[rgb]{.95,.95,.95}\textbf{91.2}&	\cellcolor[rgb]{.95,.95,.95}\textbf{95.9}&	\cellcolor[rgb]{.95,.95,.95}\textbf{97.7} & \cellcolor[rgb]{.95,.95,.95}98.6\\
\bottomrule
\end{tabular}
\caption{
AUC scores of rotation errors after the full global SfM pipeline~\cite{sweeney2015theia} averaged over all scenes of the ETH3D dataset~\cite{schops2017multi}.}
\label{tab:fp_eth3d}
\end{table}

The rotations in two reconstructions are not directly comparable since the global rotation of the view-graph is unknown.
Therefore, we first align the reconstructed and the ground truth graphs with rotation
\begin{equation}
    \hat{R}_{\text{align}} = \argmin_{R_{\text{align}}}\sum_{i=1}^{i=|\mathcal{V}|}\rho\left(\|L_{e}(R_i,R_i',R_{\text{align}})\|^2\right).
\end{equation}
The loss function $\rho(\cdot)$ we used here is Cauchy loss.
We do not use MAGSAC here to avoid bias when comparing methods to the ground truth.
Rotations $R_i$ and $R_i'$ are the orientations of view $i$ in GT and reconstructed graph, respectively. 
Finally, we rotate the reconstructed graph by $\hat{R}_{\text{align}}$.

\paragraph{Uncertainties in Rotation Averaging.}

To explore the influence of considering the uncertainties of two-view geometry estimation in rotation averaging, we compare the following weighting schemes.
The \textit{baseline} is using a constant weight for all rotations.
We test using the \textit{inlier number} of the estimated relative pose as weight in the optimization~\cite{gao2021incremental}. 
We use the proposed \textit{uncertainty}-based weighting with covariance matrices as described in \cref{sec:rot_uncertainty}.
To show its impact on multiple configurations, we run rotation averaging with Soft L$_{1}$~\cite{charbonnier1997deterministic} and MAGSAC losses~\cite{barath2021marginalizing}.

\begin{table}[htbp] 
\centering
\begin{tabular}{ r | c c c c }
\toprule
\multicolumn{1}{c}{\multirow{2}*{Weight}} 
&\multicolumn{4}{c}{AUC $\alpha$ (\%)  $\uparrow$}\\
 \cmidrule(lr){2-5}
\multicolumn{1}{c}{} & $\alpha$=2\degree & $\alpha$=5\degree & $\alpha$=10\degree & $\alpha$=20\degree\\
\midrule
Covariance & \textbf{38.3} & \textbf{62.5} & \textbf{75.8} & \textbf{84.9}\\
Trace & 38.1 &58.8 &70.9 &79.7 \\
$F$-norm & 37.5 &58.2 &70.4 &79.4 \\
\bottomrule
\end{tabular}
\caption{
Area Under the recall Curve (AUC) of the errors of the rotation averaging, on the 1DSfM dataset, when using the trace, the F-norm of the inverse covariance and proposed approach (Covariance) for weighting the view-graph edges.
}
\label{tab:uncertainty_representaion}
\end{table}

\begin{table}[htbp] 
\centering
\begin{tabular}{ r | c c c c }
\toprule
\multicolumn{1}{c}{\multirow{2}*{Loss Function}} 
&\multicolumn{4}{c}{AUC $\alpha$ (\%)  $\uparrow$}\\
 \cmidrule(lr){2-5}
\multicolumn{1}{c}{} & $\alpha$=2\degree & $\alpha$=5\degree & $\alpha$=10\degree & $\alpha$=20\degree\\
\midrule
MAGSAC~\cite{barath2021marginalizing}& \textbf{35.9} & \textbf{60.9} & \textbf{75.0} & 84.0\\
Soft L$_{1}$~\cite{charbonnier1997deterministic}& 30.4 & 54.6 & 72.0 & 80.8\\
L$_{0.5}$~\cite{chatterjee2017robust}    & 29.8 & 55.2 & 71.3 & 81.8\\
Tukey~\cite{maronna2019robust}        & 29.0 & 56.2 & 73.7 & \textbf{84.7}\\ 
Cauchy~\cite{black1996robust}       & 28.9 & 53.7 & 70.3 & 81.4\\
Huber~\cite{huber1992robust} & 21.6 & 45.7 & 63.4 & 76.7\\
GM~\cite{geman1987statistical} &13.7 & 42.2 & 65.3	& 79.0\\
Trivial       & 11.2 & 30.2 & 49.2 & 66.1\\
\bottomrule
\end{tabular}
\caption{Area Under the recall Curve (AUC) of the rotation errors of different robust losses inside the rotation averaging of~\cite{chatterjee2013efficient} averaged over all scenes in the 1DSfM dataset~\cite{wilson_eccv2014_1dsfm}.}
\label{tab:loss_func}
\end{table}

\begin{table}[htbp] 
\centering
\begin{tabular}{c |c c c}
\toprule
Downsample & \multirow{2}*{Baseline} & \multirow{2}*{Ours} & \multirow{2}*{Improvement} \\
Factor\\
\midrule
1 & 88.61 & 91.19 & +2.58 \\
4 & 81.23 & 84.74 & +3.51 \\
8 & 80.69 & 84.62 & +3.93 \\
\bottomrule
\end{tabular}
\caption{
The AUC@2$\degree$ scores of the rot.\ errors of global SfM~\cite{sweeney2015theia} on the ETH3D~\cite{schops2017multi} dataset when the images are downsampled.
}
\label{table:resolution_eth3d}
\end{table}

In \cref{tab:ra_1dsfm}, the Area Under the recall Curve (AUC) at 5$\degree$ is reported on the scenes of the 1DSfM dataset. The last row shows the average AUC scores. 
On average, using any of the compared weighting strategies improves the accuracy.
Using the proposed uncertainties leads to the highest AUC score with both robust losses.
MAGSAC clearly leads to better results than Soft L$_1$. 
The absolute best is obtained by MAGSAC loss and the proposed covariance-based weighting. 
Compared to the original Theia code (\ie, Baseline with Soft L$_1$), the proposed algorithm leads to a more than $10$ points increase in the AUC score.

\begin{table*}[htbp] 
\centering
\begin{tabular}{ c c| c c c c c c}
\toprule
\multicolumn{2}{c}{\multirow{2}*{Setting}}
& Ori. Err& Pos. Err & \# Reconst.& \# Common & Time & Reproj. Err\\ 
\multicolumn{2}{c}{}& Med (degree) & Med (meter) & Views & Views & (minute) & Avg (pixel)\\ 
\midrule
\multicolumn{2}{c|}{COLMAP~\cite{schonberger2016structure}} & - & - &14097 &-&2852&\\
\midrule
\multirow{3}*{Soft L$_{1}$~\cite{charbonnier1997deterministic}}
&Baseline & 1.44 & 1.63 & 9413 & 8824 & 27 & 1.24\\
&+ Inliers & 1.72 & 1.65 & 9398 & 8806 & 34 & 1.21\\
&+ Covariance & 1.45 & 1.60 & 9191 & 8668 & 29 & 1.18\\
\midrule
\multirow{3}*{MAGSAC~\cite{barath2021marginalizing}}
& Baseline & 1.33 & 1.72 & 9404 & 8814 & 31 & 1.04\\
&+ Inliers & 1.18 & 1.79 & 9424& 8797 & 55 & 1.21\\ 
&+ Covariance & 1.00 & 1.64 & 9361& 8767 & 46 & 0.99\\
\bottomrule
\end{tabular}
\caption{Performance of different settings for full SfM pipeline. 
We report the median orientation (in degrees) and position errors (in meters), the number of reconstructed views and views common with the ground truth, the processing time (in minutes), and the reprojection average error (in pixels). 
The COLMAP reconstruction (with default parameters) is considered ground truth.
Time is the total running time of all scenes in 1DSfM dataset without the time of the feature extraction and matching. It is measured on a laptop with Intel i7-12700H. }
\label{tab:fp-perf}
\end{table*}

The left plot of \cref{fig:cdfs} shows the rotation error distribution over all the scenes of the 1DSfM dataset. The MAGSAC loss with covariance method outperforms all other methods in all error ranges. Soft L$_1$ loss with covariance performs well when the error is smaller than 5\degree, but not as good as the Soft L$_1$ loss baseline when error get larger. One reason is that there are still some outlier edges in the view graph with small uncertainty, in this case, the optimization process would be dragged to the sub-optimal results because of the lack of robustness of the Soft L$_1$ loss.

We also show the AUCs of rot.\ errors after the full global SfM pipeline. 
As shown in \cref{tab:fp_1dsfm}, MAGSAC with covariance is still the best. 
The strategy weighting by the inlier numbers is worse than the baseline with both losses. This means that in a large-scale dataset like 1DSfM, the final bundle adjustment step could make up the gap of the information provided by the inlier numbers, which means that a stronger uncertainty extraction method (covariance) is necessary to improve the reconstruction quality.

The results of ETH3D dataset are shown in the right plot of \cref{fig:cdfs} and in \cref{tab:fp_eth3d}. 
The combination of MAGSAC and the proposed uncertainties leads to the most accurate results.
Another interesting observation is that our global SfM outperforms COLMAP method, confirming the potential in global SfMs over their incremental counterparts.  

\cref{tab:uncertainty_representaion} shows the results of rotation averaging by different uncertainty-based weighting strategies. 
We compare using the trace and Frobenius-norm of the covariance matrix to using it directly.
The direct strategy, in \cref{eq:weighted_rot_avg}, clearly outperforms both other tested algorithms. 

\paragraph{Robust Loss Functions.}

We compared the impact of using different loss functions for non-linear optimization in rotation averaging. 
In \cref{tab:loss_func}, we compare the MAGSAC loss function with other 7 popular loss functions on the scenes from the 1DSfM dataset. 
We report the AUC scores at $2^\circ$, $5^\circ$, $10^\circ$, and $20^\circ$.
MAGSAC is the best by a large margin on AUC at $2^\circ$, $5^\circ$ and $10^\circ$.
It is second best, in terms of AUC $20^\circ$, being slightly behind Tukey weighting by $0.7$ AUC points.
For example, the AUC@$2^\circ$ of MAGSAC is $5.5$ points higher that of the second best, \ie, Soft L$_1$. 



\paragraph{Image Downsampling.}
We test the reconstruction on the ETH3D dataset by downsampling the input high resolution images.
The AUC $2^\circ$ scores are reported in \cref{table:resolution_eth3d} at different downsampling factors. 
We observe that the proposed pipeline is more accurate than the baseline in \textit{all} cases.
Actually, the larger the downsampling factor (\ie, image quality reduction), the bigger the improvement compared to the baseline~\cite{sweeney2015theia}. 
This clearly demonstrates the potential in using the uncertainties of the estimated two-view geometries and the importance of choosing the best loss function. 


\paragraph{Full Global SfM Pipeline Results.}
In \cref{tab:fp-perf}, the results after the full global SfM pipeline (finishing with a bundle adjustment) are shown on the 1DSfM dataset. 
The lowest median orientation error is achieved by the proposed algorithm. 
The median position errors, the number of reconstructed views, and the number of views that are common with the ground truth, are similar in case of all methods. 


Since the complexity of evaluating MAGSAC loss is marginally larger than that of the L$_{1}$ loss and, also, we need to invert and decompose the covariance matrix, the proposed algorithm is slightly (\ie, by a few minutes) slower than other combinations. 
The benefit from global approaches is clear in this figure as their run-time is \textit{two order-of-magnitude} lower than that of COLMAP.
COLMAP runs for $48$ hours.
Global approaches run for $30$ minutes. 

\section{Conclusion}

    In this paper, we revisit the rotation averaging problem applied in global Structure-from-Motion pipelines by leveraging uncertainties in two-view epipolar geometries and investigating robust losses.
    Our experiments demonstrate that integrating the covariance matrices of the uncertainties directly in the optimization procedure largely improves reconstruction quality.
    Moreover, carefully choosing the employed robust loss also gives a boost in the accuracy.
    We believe our work helps in closing the accuracy gap between incremental and global approaches.
    The source code will be made publicly available.
    

{\small
\bibliographystyle{ieee_fullname}
\bibliography{egbib}
}

\end{document}